\documentclass{article}
\usepackage[utf8]{inputenc}
\usepackage{graphicx}
\usepackage{siunitx}
\usepackage{physics}
\usepackage{amsmath}
\usepackage{systeme}
\usepackage{subfigure}
\usepackage[dvipsnames]{xcolor}
\usepackage{array}
\newcolumntype{P}[1]{>{\centering\arraybackslash}p{#1}}
\newcolumntype{M}[1]{>{\centering\arraybackslash}m{#1}}
\usepackage{latexsym}
\usepackage{appendix}
\usepackage{amssymb}
\usepackage{lastpage}
\usepackage{fancyhdr}
\usepackage[
top    = 2.0cm,
bottom = 2.0cm,
left   = 3.0cm,
right  = 3.0cm]{geometry}
\usepackage{lineno}
\newcommand*{\QED}{\null\nobreak\hfill\ensuremath{\square}}%

\title{Event Prediction and Causality Inference\\Despite Incomplete Information\\ \vspace{0.3cm}- An Analytical, Simulation and ML Approach -}
\author{\vspace{-0.5cm}Harrison Lam, Yuanjie Chen, Noboru Kanazawa, Mohammad Chowdhury,\\
Anna Battista, and Stephan Waldert}

\fancypagestyle{firststyle}
{
   \fancyhf{}
   \fancyfoot[L]{Page \thepage\ of \pageref{LastPage}}
}

\begin{document}
\renewcommand{\arraystretch}{2.5}
\renewcommand{\headrulewidth}{0pt}

\maketitle
\thispagestyle{firststyle}

\begin{abstract}
We explored the challenge of predicting and explaining the occurrence of events within sequences of data points. Our focus was particularly on scenarios in which unknown triggers causing the occurrence of events may consist of non-consecutive, masked, noisy data points. This scenario is akin to an agent tasked with learning to predict and explain the occurrence of events without understanding the underlying processes or having access to crucial information. Such scenarios are encountered across various fields, such as genomics, hardware and software verification, and financial time series prediction.\\
We combined analytical, simulation, and machine learning (ML) approaches to investigate, quantify, and provide solutions to this challenge. We deduced and validated equations generally applicable to any variation of the underlying challenge. Using these equations, we (1) described how the level of complexity changes with various parameters (e.g., number of apparent and hidden states, trigger length, confidence, etc.) and (2) quantified the data needed to successfully train an ML model.\\
We then (3) proved our ML solution learns and subsequently identifies unknown triggers and predicts the occurrence of events. If the complexity of the challenge is too high, our ML solution can identify trigger candidates to be used to interactively probe the system under investigation to determine the true trigger in a way considerably more efficient than brute force methods.\\
By sharing our findings, we aim to assist others grappling with similar challenges, enabling estimates on the complexity of their problem, the data required and a solution to solve it.
\end{abstract}
\vfill
\textcolor{gray}{
For any inquiries on this paper, please contact SW, HL and YC at science@waldert.de, Harrison7871@hotmail.com and chenyuanjie625@hotmail.com, respectively.\vspace{0.3cm} \\
CC BY-NC-SA 4.0 Attribution-NonCommercial-ShareAlike 4.0 International}

\newpage

\fancyfoot[L]{Page \thepage\ of \pageref{LastPage}, Lam et al. 2024, Event Prediction and Causality Inference}

\section{Methodology}\label{problem_statement_sec}
\subsection{Problem formulation and terminology}
Consider an arbitrary sequence $X=[x_1,x_2, x_3,\dots]$ followed by the occurrence of an event $E$, i.e. $X\Rightarrow E$. We want to automatically infer what in $X$ caused the occurrence of $E$.\\
The challenge is that each element $x_i$ has an apparent state $A=\left\{1,2,\dots,a\right\}$ and a hidden state $H=\left\{1,2,\dots,h\right\}$, with $(a,h,i) \in \mathbb{N}$ and states independently and uniformly distributed. If we now look at a sub-sequence of length $n$ in $X$ and directly preceeding $E$, the total number of possible apparent sequences is $a^n$, however, because of the hidden states, the total number of actually possible (but partly non-observable) sequences is $(ah)^n$.\\
Moreover, the event $E$ is triggered deterministically by a trigger $t\in T$ within $X$, e.g. $[x_2,x_3,x_5]\Rightarrow E$. We refer to $T$ as the set of triggers causing the occurrence of event $E$. While each trigger $t$ is ordered and has to occur before $E$ (causality), the trigger can occur at any distance before $E$, can be of any length $l\leq n$ and can consist of consecutive (e.g. $[x_3,x_4,x_5]$) or non-consecutive (e.g. $[x_3,x_8]$) elements in $X$. Please note that any particular event $E$ only occurs if the trigger consists of elements of particular apparent and, importantly, particular hidden states.\\
We use $*$ to represent any elements and any number of elements. For example, we write $[x_i*x_n]$ for a sub-sequence of any length starting with $x_i$ and ending with $x_n$ that can contain consecutive or non-consecutive elements separated by any number of (trigger-irrelevant) elements.

\subsection{Problem statement}
We assume no information is available about the underlying process of element occurrence and event generation, hence, we assume apparent and hidden states occurring independently and randomly following a uniform distribution.\\
We have encountered such challenges in our sequential data and needed to understand what causes an event in order to optimise workflows and improve processes. Moreover, we wanted to understand the complexity of the underlying problem, which training data size is required and which length should be chosen for the 'search window' to make it likely for machine learning to be able to successfully infer the trigger. Hence, we state the following problems:\\
\\
\textit{Problem 1}: Depending on $a$, $h$ and $l$, what is the relation of $n$ and the probability of being able to find $t$, any trigger that triggers the event, i.e. what is the probability of $t$ existing in the window of length $n$?\\
\textit{Problem 2}: Without knowing $l$ of any $t$ and with $x_i$ having non-observable states: infer $T$ from $X$.\\
\textit{Problem 3}: What is the data size required to address \textit{Problem 1} and \textit{Problem 2}?

\subsection{Problem example}\label{problem_example_sec}
Let us consider a trivial example in which each element can take on two apparent states ($a=2$): Leave ($L$) and Stay ($S$), four hidden states ($h=4$) and the trigger length is three ($l=3$).\\
A possible apparent sequence of $n=10$ may be $[LSLLLSSLSS]\Rightarrow E$, with the actual but non-observable sequence being $[L_HS_HL_HL_HL_HS_HS_HL_HS_HS_H]\Rightarrow E$, i.e. although the apparent sequence is known, there are $h^n=1,048,576$ possible sequences.\\
Furthermore, there are $a^l=8$ possible unique apparent triggers (e.g. one is $[LLS]$) and $(ah)^l=512$ possible unique actual triggers (e.g. one is $[L_1L_1S_1]$). \\
We would now want to solve \textit{Problem 2}, using information of \textit{Problem 1} to deduce which length of the search window $n$ needs to be chosen to find $t$ with a probability of, for example, 95\%.

\subsection{Problem solving approaches}
We used three approaches to understand and solve these problems: analytics, simulation and machine learning.\\
In Section \ref{Analytical_section}, we used the first two approaches to derive and numerically prove equations allowing to directly quantify the complexity of the problem and the required data size to find $T$, leading to answers of \textit{Problem 1} and \textit{Problem 3}. We start with a simplified version of the problem, which we then generalise to derive equations for any trigger length and number of apparent and hidden states. These equations then also provide solutions to the window length and required data size, i.e. how much data need to be gathered and analysed to be able to find the real trigger.\\
In Section \ref{ml_section}, we outlined a machine learning approach to provide a solution for \textit{Problem 2}, i.e. automatically extracting the trigger causing the event and, hence, an understanding of the underlying process driving event occurrence.  

\section{Analytics and simulation: window and data size}\label{Analytical_section}
Once we observe events and want to infer what caused their occurrence, we want to estimate the size of the search window we should apply and how much data we need in order to gather 'sufficient' information.\\
Our first aim is to answer \textit{Problem 1}: Depending on $a$, $h$ and $l$, what is the relation of $n$ and the probability of being able to find $t$ in a sequence of length $n$, or alternatively: what is the probability of $t$ existing in the window of length $n$? We then answer \textit{Problem 3}. 
\subsection{Simplified problem}
We start with a set of fixed parameters and will then generalise. As a reminder, let us consider two apparent states ($a = 2$): Leave ($L$) and Stay ($S$), four hidden states ($h = 4$) and a trigger length of three ($l = 3$). The real but unknown trigger is, for example, $t=[L_1*L_3*S_2]$  and a possible sub-sequence could be $[L_2L_4L_1S_2L_3L_3S_2E]$.
\subsubsection{Simplified problem: chances of any particular trigger existing}
As we assumed randomly, independently and uniformly distributed states, the probabilities $p$ of any particular trigger of length $l=3$ to occur in a sequence of length $n\leq 3$ are
\begin{equation}\label{eq:1}
    p(t|n=0)=p(t|n=1)=p(t|n=2)=0, \quad n<3
\end{equation}
\begin{equation}\label{eq:2}
    p(t|n=3,a=2,h=4,l=3)=\frac{1}{(2*4)^3}=\frac{1}{512}<0.002, \quad n=3
\end{equation}
\\
In general, for sequence lengths $n\geq 3$, the probability of a particular trigger of length three to occur by chance can be described using binomial equations: 

\begin{equation}\label{eq:3}
    p(t|n,a=2,h=4,l=3)=
    \sum_{k=3}^n\binom{n}{k}\left(\frac{7}{8}\right)^{n-k}\left(\frac{1}{8}\right)^k, \quad n\geq 3
\end{equation}

\subsubsection{Simplified problem: only triggers with elements of identical hidden states cause events}
There is a sub-class of the general problem: events are caused exclusively by triggers composed of elements with the same hidden state. This scenario is not unlikely to be encountered in real-life applications. If the underlying process is known to have this feature, we can use $P$ (see Equation \ref{eq:4}, instead of $p$) to answer \textit{Problem 1}.\\
Still following the simplified example, we now know that the trigger can only be of the following form: $t=[L_i*L_i*S_i]$ with $i\in H$.\\
Since $p(n)$ for any hidden state is equivalent, we can compute $P(n)$, the probability that triggers from either of the four hidden states ($h=4$) occur in a sequence of size $n$:

\begin{equation}\label{eq:4}
\begin{aligned}
    P(n)
    &= p(t=L_i*L_i*S_i|n,a=2,h=4) \\
    &= 1 - p(\lnot \, (L_1*L_1*S_1 \,or\, L_2*L_2*S_2 \,or\, L_3*L_3*S_3\, or \, L_4*L_4*S_4)|n,a=2,h=4) \\
    &\gtrapprox 1 - \prod_{i=1}^4 p(\lnot L_i*L_i*S_i|n,a=2,h=4) \\
    &= 1 - (1-p(t=L_i*L_i*S_i|n,a=2,h=4))^4 \\
    &= P_t(n)
\end{aligned}
\end{equation}
The $\gtrapprox$ is used due to the fact that $p(\lnot L_i*L_i*S_i|n,a,h)$ are not truly independent for $i\in H $, this is due to the fact that we are limiting the number of possible outcomes. For example, 
\begin{equation}\label{eq:5}
    P(\lnot L_1L_1S_1 | \lnot (L_i*L_i*L_i) , 2\leq i\leq 4,n,a=2,h=4) \lessapprox P(\lnot L_1L_1S_1|n,a=2,h=4)
\end{equation}
$P_t(n)$ provides a slight underestimation of the probability $P(n)$ and we can summarise
\begin{equation}\label{eq:6}
    P(n)\gtrapprox P_t(n)=1-(1-p(n))^4
\end{equation}
Our simulations (see Section  \ref{Simp_Simulations}) also confirmed this relationship and we found the difference to be small, but observable (Figure 1). For the remainder of this manuscript, we chose to argue without repeatedly referring to this point, i.e. although quantitatively existing, we qualitatively ignore this difference as it does not impact our findings and conclusions. 

\subsubsection{Simplified problem: simulations}\label{Simp_Simulations}
\begin{figure}[h]
\centering
\begin{subfigure}
  \centering
  \includegraphics[width=.49\linewidth]{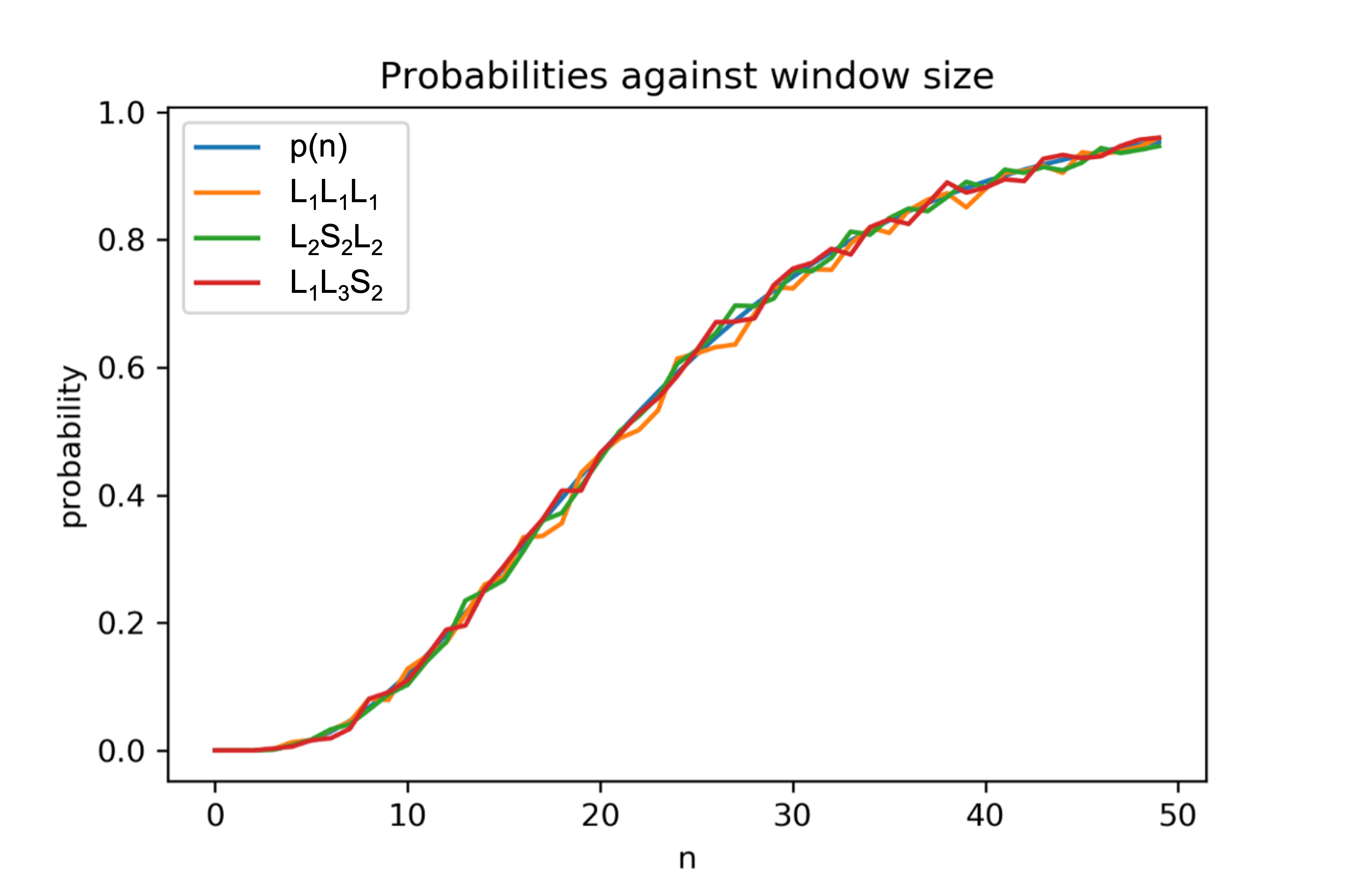}
\end{subfigure}%
\begin{subfigure}
  \centering
  \includegraphics[width=.49\linewidth]{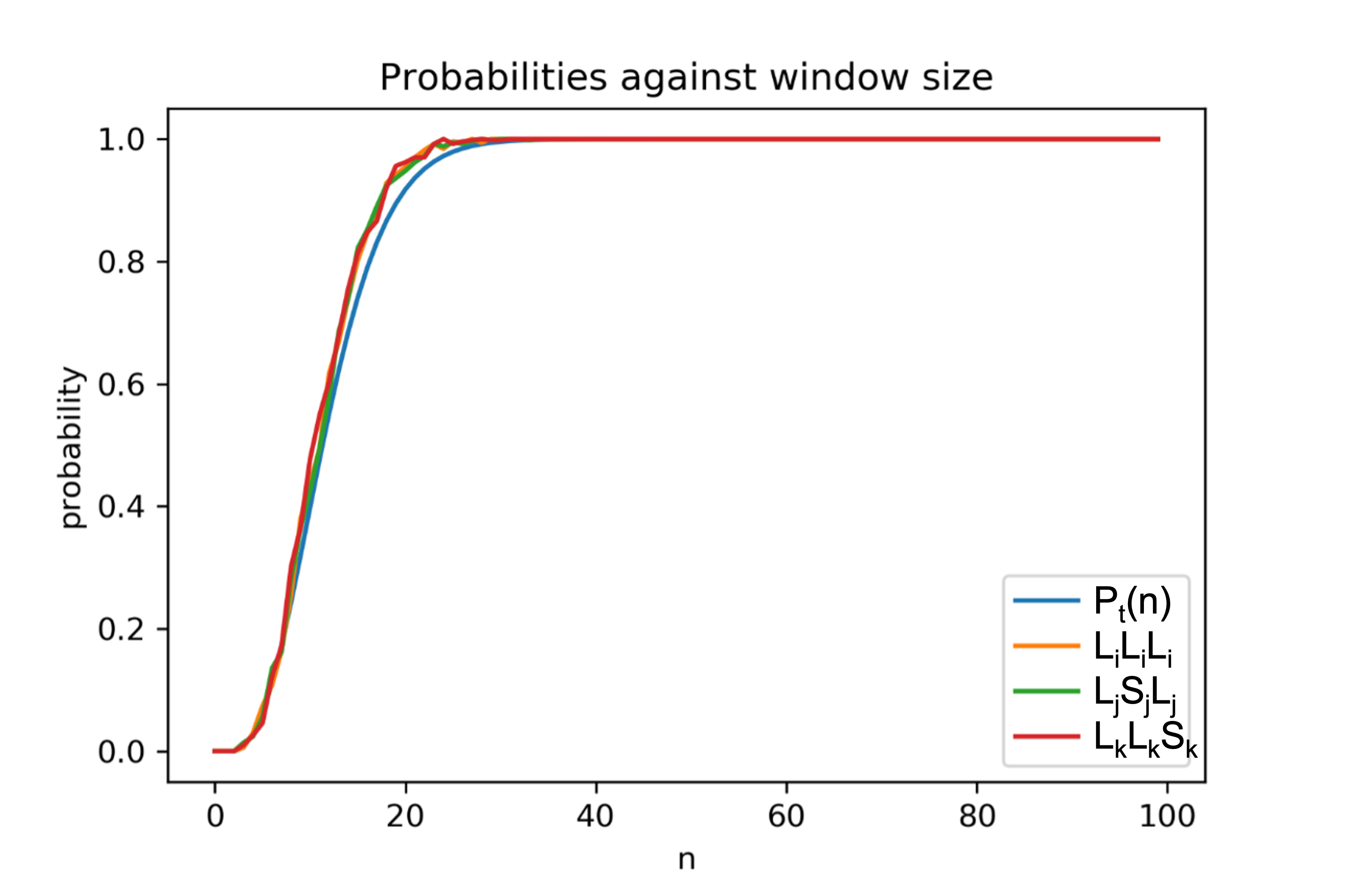}
\end{subfigure}
\caption{Left: $p(n)$ (any apparent state and any particular hidden state) vs simulations: numerical confirmation of the correctness of $p(n)$ and that simulations return the same results for different types of triggers. Right: $P(n)$ (any apparent state and any hidden but same state) vs simulations: numerical confirmation that $P_t(n)$ can
serve as an estimate as well as lower bound of $P(n)$. The observation is consistent with Equation \ref{eq:4}.}
\label{fig:s1}
\end{figure}

To assess Equations \ref{eq:3} and \ref{eq:4}, for sequence lengths between 0 to 50, we run 1000 simulations in which we randomly generated apparent states, each with a hidden state. We then calculated the probability that a (random) trigger is contained in a simulated sequence by dividing the number of apparent triggers by 1000. In the left plot in Figure \ref{fig:s1}, we numerically confirmed that $p(n)$ is the probability of the presence of any particular trigger in a sequence of length $n$, and that there is no difference among different types of trigger (e.g. LLL vs LSL), as expected.
\\In the right plot in Figure \ref{fig:s1}, we numerically confirmed that we can use $P_t(n)$ as an estimator as well as lower bound of $P(n)$. This is consistent with Equation \ref{eq:4} and Equation \ref{eq:6}.  %TODO

\begin{figure}
    \centering
    \includegraphics[width=.7\textwidth]{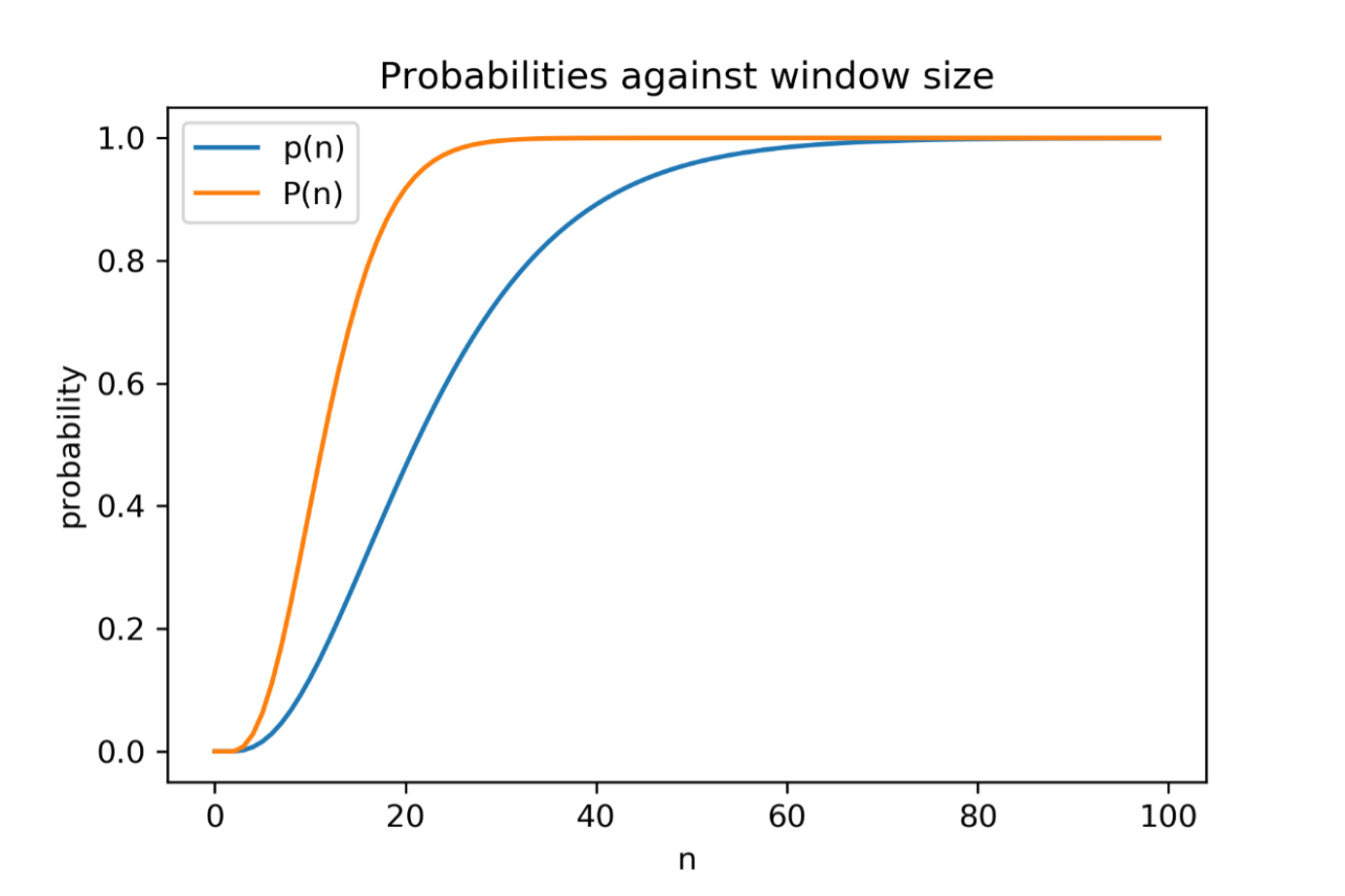}
    \caption{The probabilities $p(n)$ and $P(n)$ against different sequence lengths. We can see that given the same sequence length, $P(n)\geq p(n)$}
    \label{p vs P}
\end{figure}

\subsubsection{Simplified problem: window length required for 95 percent confidence of containing a trigger}\label{95confidence}
For both $p(n)$ and $P(n)$, we can determine a window length $n$ so that a trigger will be contained within this window at a certain confidence, and vice versa.\\
For example, using the simplified problem ($a=2,h=4,l=3$), a sequence length of $n=22$ or $n=60$ is needed for us to be $95\%$ confident for a trigger of type $p(n)$ or $P(n)$, respectively, to be contained in this window (see Figure \ref{p vs P}). 

\subsubsection{Simplified problem: minimum data set needed to infer the true trigger}\label{min_data_subsec}
\begin{figure}[h]
\centering
\begin{subfigure}
  \centering
  \includegraphics[width=.51\linewidth]{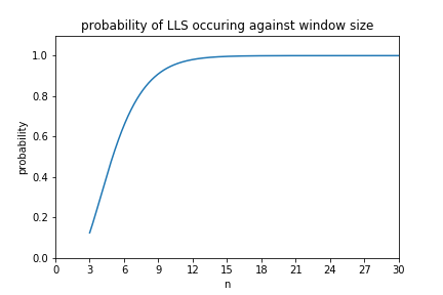}
\end{subfigure}%
\begin{subfigure}
  \centering
  \includegraphics[width=.46\linewidth]{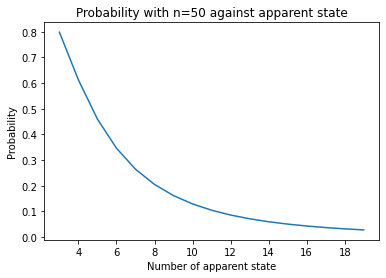}
\end{subfigure}
\caption{Figure on the left shows the probability of $'LLS'$ occurring in a random sequence Q plotted against $n$. For the figure on the right, we have used the generalised formed and fixed the window length $n=50$, $h=4$ and $l=3$. As we can see, even with a large window length, the sequence quickly becomes very difficult to solve.}
\label{fig:f3}
\end{figure}

We have discussed problems \textit{Problem 2} and \textit{Problem 1} for the simplified case. We remain with the simplified case and address \textit{Problem 3} now: What is the data size required to infer the trigger? We look at the sequence as in our real-world scenario, i.e. we have no information about the hidden states.
For the simplified case ($a=2$, $h=4$, $l=3$), there are $a^n$ different, individual apparent sequences of length $n$. The objective here is to determine how many of these sequences contain, for example, $'L*L*S'$ (by chance).\\\\
\textbf{Corollary:} There are $\frac{n^2+n+2}{2}$ sequences of length $n$ not containing the trigger, e.g. $'L*L*S'$.\\\\
\textit{Informal Proof}: \\%Proof that needs to be worked on
Akin to a Galton board, arrange all possible sequences of length $n$ in a pyramid-like notation: start with the $L$-only sequence and in every subsequent line, list all combinations with one more $S$ and one less $L$ than in the previous line, until arriving at the $S$-only sequence:\\
\begin{align*}
    &LLL\dots LLL, &line\: 1\\
    &SLL\dots LLL,\, LSLL\dots LLL,\, \dots,\, LLL\dots LLS, &line\: 2\\
    &\dots, \\
    &LSS\dots SSS,\, SLSS\dots SSS,\, \dots,\, SSS\dots SSL, &line\: n\\
    &SSS\dots SSS &line\: n+1
\end{align*}
We know:
\begin{itemize}
    \item There are $n+1$ lines.
    \item The number of $S$ in the sequences in line $n$ equals $n-1$.
    \item Last two lines do not contain $'L*L*S'$ because they have only one or no $'L'$.
    \item In the $k^{th}$ line, where $1 \leq k \leq n$,  there are $k$ sequences not containing $'L*L*S'$.
    \\
    Reason: In the sequences in the $k^{th}$ line, the number of $S$ and $L$ is $k-1$ and $n-k+1$, respectively. To not have $'L*L*S'$, $n-k$ lots of $L$ have to appear at the end of the sequence. The remaining one $L$ can appear at any of the $n-(n-k)=k$ positions before. Hence, there are $k$ sequences in the $k^{th}$ line not containing $'L*L*S'$.
    \item The last line contains 1 sequence and does not contain $'L*L*S'$.
\end{itemize}
Based on these observations, we can calculate the overall number of sequences not containing $'L*L*S'$ by summing across the number of sequences not containing $'L*L*S'$ per line:
\begin{equation}\label{eq:7}
    \left(\sum_{k=1}^{n}k\right)+1=\frac{n(n+1)}{2}+1=\frac{n^2+n+2}{2}
\end{equation}\QED
\\\\
It follows that the number of sequences (out of the total number of sequence $(=2^{n})$) containing the target sequence is
\begin{equation}\label{eq:8}
    N(n)=2^n-\frac{n^2+n+2}{2}
\end{equation}
and the probability of $'L*L*S'$ occurring in a random sequence of length of $n$ is
\begin{equation}\label{eq:Q}
    Q(n)=\frac{N(n)}{2^n}=1-\frac{n^2 + n + 2}{2^{n+1}}, \quad n>2
\end{equation}
which increases quickly with large sequence lengths (Figure \ref{fig:f3}).\\\\
We now come back to the question: what's the minimum data size required to infer the true trigger (see \textit{Problem 3}).\\
As we have shown before (Section \ref{95confidence}), for a $95\%$ confidence that the true trigger is contained in a sequence, we require the window length to be a minimum of 22. See Figure \ref{fig:f3}.\\
Using this value in equation \ref{eq:Q}, we calculate a probability of $'L*L*S'$ occurring in a random sequence of length $n=22$ by chance as $Q(22)=0.999939441681$. This is independent of hidden states.\\
To be able to irrevocably exclude a particular trigger despite not knowing its hidden states, we want to observe at least one sequence without this particular trigger to cause an event. Then we know that the event must have been caused by a different trigger, a process akin to eliminating false trigger candidates. What is the data size required enable this observation (\textit{Problem 3}).\\
Hence, for a trigger length of $l=3$, the probability to be able to exclude one of the $a^{l}=8$ triggers in a single sequence is $1-Q(22)$. This probability is very low. The probability of a particular sequence to not enable us to eliminate a false trigger candidate is $1-(1-Q(22))=Q(22)=G$ (please note: mathematically $Q$ and $G$ are the same, however, the interpretation has changed here).\\
Now, if we want to be able to eliminate a single trigger candidate at higher chance level than 95\%, what is the number of sequences (data size) required? In other words, we want to derive how many sequences $m$ are required such that $G^{m}<0.05$. 
Solving this equation with $G=Q(22)=0.99993944168$ results in $m>49467$, i.e. we require this number of sequences to be able to exclude one of the false triggers.\\
Please note that for simplicity, we do not consider scenarios in which these sequences allow to exclude more than one false trigger. Therefore, the following arguments and numbers depict an upper bound on the complexity.\\
In our thought experiment, we aim to exclude one particular false trigger out of 7 false trigger candidates, which is different to excluding any of the 7 false trigger candidates. Here we use the initial runs to exclude 6 out of the 7 false trigger candidates (repeatedly), while aiming to eliminate the remaining last false trigger. As in every run of observing 49467 sequences we hit any of the 6 false triggers with a probability of $\frac{6}{7}$, how many times ($r$) do we then need to repeat the run to reduce the probability of not eliminating one of the 6 false triggers, but actually the 1 remaining false trigger, to below 0.05, hence, $\left(\frac{6}{7}\right)^r<0.05$? \\
Solving this equations returns $r=20$, i.e. we need to observe 20 times 49467 sequences to reduce the probability of only eliminating 6 instead of all 7 false triggers to $<0.05$, or vice versa, we require that many sequences to be able to exclude even the last remaining false trigger with a probability of 95\% or higher. That means we require $m*r=989341$ sequences to eliminate all false triggers with a probability of 95\% or higher.\\
Therefore, for non-consecutive trigger, to be able to identify the true trigger with a probability of 95\% or higher, we require $m*r=989341$ sequences of length $n=22$. This is the data size required and also the answer to \textit{Problem 3}.

\subsection{Trivial case: consecutive trigger}
We mention consecutive trigger, where elements of a trigger are consecutive in the sequence, only briefly here. This is a further simplification and simply a subset of the challenge. The consecutive trigger could come in two forms: Consecutive trigger that leads directly to the event (scenario 1) and consecutive trigger that is embedded in the sequence with the event occurring sometime later (scenario 2). For both scenarios, if we know triggers are of type 'consecutive', identification of triggers is trivial and only requires small data sets (without proof).

\subsection{Generalised problem}
For clarity and visualisations, we have so far only argued on the basis of the simplified problem ($a=2$, $h=4$ and $l=3$). In Table \ref{table:gen}, we generalised $p(n)$ for any number of hidden and observable states and any trigger length.\\
Equations $p_2(n)$, $p_3(n)$ and $p_4(n)$ in Table \ref{table:gen}, are equivalent to $p(n)$ but were derived differently (please see Appendix).\\
The generalised formula for $P(n)$ can be written as $P^g(n)\gtrapprox 1-(1-p^g(n))^h$. Where $p^g(n)$ is any of the three generalised $p^g(n)$, or $p_{2}^g(n)$ or $p_{3}^g(n)$.

\begin{table}[h]
\footnotesize
\hskip-0.05cm\begin{tabular}{|p{7.3cm}|p{7.5cm}|}
    \hline
    \textbf{Simplified Problem $(a=2, h=4, l=3)$} & \textbf{Generalised Problem $(a, h, l)$} \\
    \hline
    \(\displaystyle p(n) =\sum_{k=3}^n\binom{n}{k}\left(\frac{7}{8}\right)^{n-k}\left(\frac{1}{8}\right)^k \quad n \geq 3\) & \(\displaystyle p^g(n) =\sum_{k=l}^n\binom{n}{k}\left(\frac{ah-1}{ah}\right)^{n-k}\left(\frac{1}{ah}\right)^k \quad n > l\) \\[0.2cm]
    \hline
    \(\displaystyle p_{2}(n)=\sum_{i=3}^n\binom{i-3}{3-1}\left(\frac{7}{8}\right)^{i-3}\left(\frac{1}{8}\right)^3\) & \(\displaystyle p_{2}^g(n)=\sum_{i=l}^n\binom{i-1}{l-1}\left(\frac{ah-1}{ah}\right)^{i-l}\left(\frac{1}{ah}\right)^l\) \\[0.2cm]
    \hline
    \(\displaystyle p_{3}(n+1)=\frac{7}{8}p_{3}(n)+\frac{1}{8}\left(1-\left(\frac{7}{8}\right)^n-\left(\frac{7}{8}\right)^{n-1}\times \frac{n}{8}\right) \) & 
    
    $\begin{aligned}
        \displaystyle p_{3}^g(n+1)&=\frac{ah-1}{ah}p_{3}^g(n)\\
        &+\frac{1}{ah}\left(1-\left(\frac{ah-1}{ah}\right)^n-\left(\frac{ah-1}{ah}\right)^{n-1}\times \frac{n}{ah}\right)
    \end{aligned}$\\
    \hline
    \(\displaystyle p_{4}(n+1, t) = \frac{7}{8}p_{4}(n,t)+\frac{1}{8}p_{4}(n,l-1)\) & \(\displaystyle p_{4}^g(n+1, l) = \frac{ah-1}{ah}p_{4}^g(n,l)+\frac{1}{ah}p_{d4}^g(n, l-1)\) \\[0.5cm]
    With initial condition: \(\displaystyle p_{4}(0,t)=0, \) &
    With initial condition: \(\displaystyle p_{4}^g(0,t)=0, \) \\
    \(\displaystyle p_{4}(n,0)=1, p_{4}(n,n)=\left(\frac{1}{8}\right)^n \)
    & \(\displaystyle p_{4}^g(n,0)=1, p_{4}^g(n,n)=\left(\frac{1}{ah}\right)^n \) \\
    \hline
\end{tabular}
\caption{Comparing equations with specific settings against generalised equation}
\label{table:gen}
\end{table}

\section{Machine learning approach for Trigger Identification}\label{ml_section}
In this section, we propose a machine learning (ML) model capable of identifying the hidden trigger sequence. We generated simulation data to demonstrate how the model extracts the trigger sequence.
\subsection{ML model selection}\label{ml_model_selection}
There are multiple models that could potentially identify the trigger \cite{1, 2}. We could  formulate this as a sequence labelling problem, which would allow us to identify the trigger directly using models such as Conditional Random Fields (CRF) \cite{3}. However, since the trigger is unknown, we would not be able to provide (supervised) labels to each element. Although there have been studies on unsupervised trigger identification for natural language \cite{4, 5}, which learn based on the positional pattern and grammatical structure of the language, we decided not to take this approach due to two reasons. Firstly, in the challenge described here, each element of the trigger sequence occurred randomly, hence transfer learning based on position will not be useful. Secondly, the performance of zero-shot methods generally remain lower than supervised methods. We have therefore formulated the challenge here as a supervised learning problem, using findings derived in Section \ref{Analytical_section} to obtain the required window size, and whether the sequence triggers an event.\\
Other approaches for this challenge have also been considered, one such approach is a Hidden Markov Model (HMM) to find the hidden trigger sequence \cite{6,7,8}. However, our challenge does not lend itself to assumptions in HMM, for example that the future state only depends on the current state. Moreover, for a sequence that is generated randomly, all transition probabilities between elements will be equivalent, which carries no information and does not contribute to identifying the trigger sequence.\\
We have chosen a deep learning architecture that is able to identify the hidden pattern in each sequence-event pair. The architecture consists of multiple embedding layers and attention layers that help identifying the trigger. Details of the architecture are explained in Section \ref{model_architecture}.

\subsection{Experiment setup}\label{ml_experiment}
To generate simulation data, we created sequences with apparent and hidden states randomly, independently and uniformly distributed. Triggers were randomly and manually defined a priory. In the simulated data, we then scanned the random sequences and inserted an event if the previously defined trigger was detected  and depending on the scenario under investigation (please see below).\\ Following our main example as introduced above, we used a trigger length of $l=3$ and $h=4$ and $a=2$. We also examined our model with trigger length of $l=5$ and $l=7$, resulting in similar findings (not presented here).\\
For $l=3$, datasets with one million random elements were produced, which corresponds approximately to the minimum sequence length required to identify the true trigger, as derived in Section \ref{min_data_subsec}.\\
Through this process, we created artificial datasets with the same characteristics as we faced in our real-world challenge, but with the ground-truth known and testable.\\
The ML model (Section \ref{model_architecture}) was trained and tested on different datasets, reflecting the following four scenarios (four types of event trigger sequences):

\begin{itemize}
    \item consecutive trigger with no hidden state
    \item consecutive trigger with hidden states
    \item non-consecutive trigger with no hidden state
    \item non-consecutive trigger with hidden states.
\end{itemize}

\begin{figure}
    \centering
    \includegraphics[width=.7\textwidth]{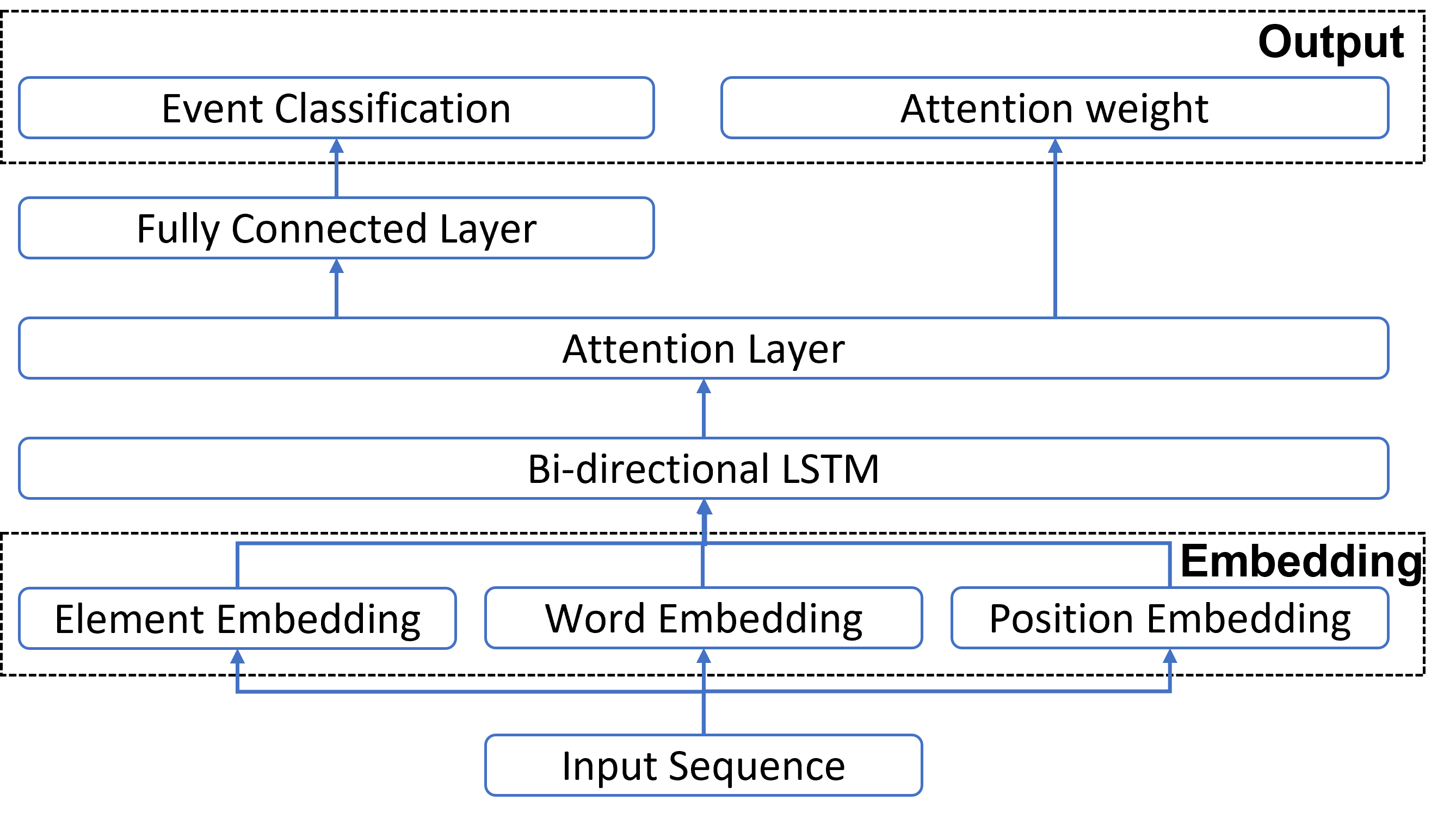}
    \caption{The model architecture for trigger identification.}
    \label{fig:architecture}
\end{figure}

\subsection{Model architecture}\label{model_architecture}
In our investigations, we conducted a series of experiments involving various deep learning architectures prior to arriving at the architectural configuration described in this paper \cite{9}. For instance, one of the preliminary iterations involved a more straightforward model featuring a single embedding layer complemented by an attention layer. Our empirical observations revealed that this model yielded sub-optimal performance, as it failed to produce accurate outputs, a deficiency that can plausibly be attributed to the lack of informative inputs supplied to the attention layer.\\The deep learning architecture we propose to solve the challenge utilises a recurrent neural network with additive attention mechanisms (Figure \ref{fig:architecture}) and the following specifications:
\begin{enumerate}
    \item \textbf{Multiple Embedding layer}: There are three different embedding layers, each extracting different information about the sequence. The first embedding layer considers the apparent state of each individual element. The second embedding layer considers all nearby elements around a position, similar to an N-gram approach. The third embedding layer is a position encoder \cite{10}. All embedding vectors are added before feeding to the next layer.
    \item \textbf{LSTM Layer}: Producing a hidden state for each embedding vector, the LSTM layer is treated as another embedding layer, which encodes the sequence into higher dimensions. Hence, all the hidden states produced by the LSTM layer are considered.
    \item \textbf{Attention Layer}: We utilised additive attention, which uses feed-forward layers to calculate the attention weight. The weight is multiplied with each hidden state produced from the LSTM layer and the output summed up to a single vector. We then retrieve the attention score from the attention layer, this will be used to find the trigger sequence.
    \item \textbf{Output layer}: The output layer is a fully-connect layer, which calculates the probability at which the sequence will trigger an event. It also returns the attention weight of each element, which is the main predictive source for trigger identification (see Section \ref{method_trigger_id}).
\end{enumerate}
This architecture was inspired by multiple LSTM architectures \cite{11,12}. Using attention weights to interpret results has been used previously \cite{13,14}. We decided to use this architecture because most of the elements in the input sequence do not contribute to the event. Hence, using the attention layer, we aimed to identify important elements that contribute to the event. With the help of multiple embedding layers, which include information about nearby elements and the position of the element, the model has a better 'understanding' of the effects of nearby elements, and their position, on the probability of causing an event. \\
To ensure the model can identify the trigger using the methodology in Section \ref{method_trigger_id}, we need the attention layer to be our key predictor on the classification problem, hence carefully tuning the hyper-parameters is an important step for success.

\subsection{Method for Trigger Identification}\label{method_trigger_id}
To train the model, the arbitrary long sequence $X$ was broken down into many sequences with window length $n=22$, as calculated from Section \ref{95confidence} for a desired 95\% confidence. We determined whether the training sequence triggers an event by looking whether the end of the sequence contained the event element $E$ that was inserted by the set rules in Section \ref{ml_experiment}. To allow the model to understand the importance of position, we have combined sequences of various lengths, all corresponding to a more than 95\% confidence level of containing the trigger sequence within the training sequences.\\
Although the model architecture reflects a classification in terms of whether a sequence under consideration will trigger an event or not, using potential trigger as the input for the model to identify the trigger has proven to be inaccurate. The inaccuracy is mainly caused by the difference of sequence length between potential triggers (length of 3) and the training sequence (length of 22), hence requiring many padding tokens, which does not contribute to the prediction of the event (because this is equivalent to giving less information to the model).\\
Therefore, we used the attention weight as the trigger identifier. For each training sequence, we looked at the top $K$ elements that have the highest attention weights to construct the trigger (with $K$ equals length of the trigger sequence). In other words, we inspected which elements the model focused on while making the prediction. \\
To give a concrete example, if we know the trigger length is 3 and the input sequence is [$L$,$L$,$S$,$L$], with output attention weight $[1.0,0.0,0.75,0.5]$, the constructed trigger sequence will be $LSL$ (first, third and fourth element).\\
After training, we tested the model on new data (not yet seen by the model) and counted the constructed sequence for each training sequence, with the highest count being the trigger sequence.

\subsection{Results of Machine Learning approach}

\begin{figure*}
   \centering
\begin{tabular}{|M{1.5cm}|M{6cm}|M{6cm}|}\hline
&\textbf{Consecutive sequence}&\textbf{Non-consecutive sequence}\\\hline
\textbf{No hidden state}&
\includegraphics[width=6cm]{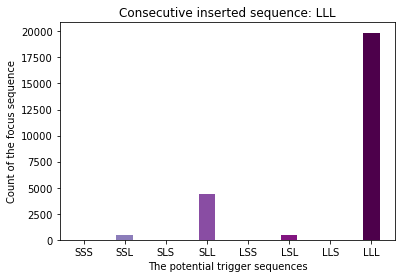}&
\includegraphics[width=6cm]{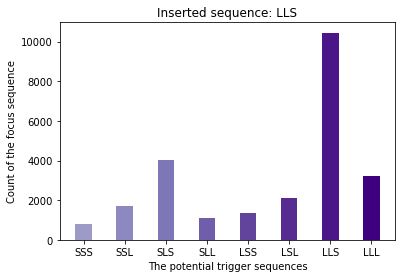} \\\hline
\textbf{With hidden state}&
\includegraphics[width=6cm]{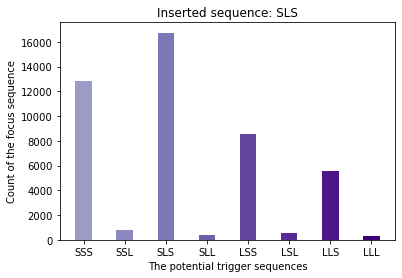}&
\includegraphics[width=6cm]{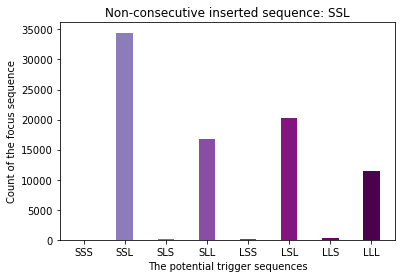}\\\hline
\end{tabular}
\caption{Results of applying the ML model to all four types of trigger sequences (scenarios in Section \ref{ml_experiment}). The y-axis shows how often the model paid the highest attention to each potential trigger sequence indicated on the x-axis. The highest counts were always obtained for the actual trigger sequence.}
    \label{fig:ML_result} % I can do without the label too
\end{figure*}

For simplicity, only results for datasets using a trigger length $l=3$ are presented here. Results for longer trigger lengths were similar, but required longer training time and larger testing set to identify the trigger. \\
After training the model on the simulated data, we tested it on unseen data for validation and obtained the attention weights for the entire dataset (sequence-event pair). For each input sequence, we detected the top three elements (trigger length of 3), those with the highest attention weight, to infer the trigger sequence (see Section \ref{method_trigger_id} for more detail).\\
For both, consecutive and non-consecutive triggers: With no hidden states (top row, Figure \ref{fig:ML_result}), the model was able to identify the actual trigger sequence with high confidence. With hidden states (bottom row, Figure \ref{fig:ML_result}), the model suggested some other potential trigger sequences (as expected), yet, the highest count was indeed identifying the actual trigger sequence.\\
In summary, the results displayed in Figure \ref{fig:ML_result} show that the highest counts were obtained for the actual trigger sequence, i.e. the model we designed was able to successfully infer the true trigger in all scenarios (hidden states - no hidden states - consecutive - non-consecutive triggers).

\section{Discussion}\label{discussion}
We investigated different scenarios of the challenge, and developed analytical and machine learning solutions, to predict events and infer causality despite crucial information being inaccessible.\\ 
We derived equations, summarised in Table \ref{table:gen}, applicable for any number of apparent states, hidden states and trigger length and type, to determine the window length so that the hidden trigger is contained at any desired confidence levels. Users can thus estimate the amount of data needed for analysis, particularly helpful for domains where data are limited. Moreover, this also solves issues where the input sequence is too long for a machine learning model to handle.\\
To find the window length, we used various methods to calculate the probability $p(n)$ (Table \ref{table:gen}). All methods followed a different logic, but returned the same results, as shown in Figure \ref{Different probability equation.} (Appendix). The equations
are applicable in scenarios without clear segmentation, variable vocabulary size (apparent state) and with or without hidden information (number of hidden states, including zero).\\
Hidden states were assumed to be discrete and their number known, which may not always be true, especially in biology, for example. However, this additional challenge might be mitigated with better or different measuring devices or solved in an iterative approach, varying the number of states in the analysis and testing for highest performance.\\
The equations do not only describe the complexity of the challenge but also enable two important steps in solving the challenge: Firstly, as mentioned above, finding the optimal window length given a self-chosen confidence level, which limits the need of analysing long input sequences, which further increases noise. Secondly, inferring the minimum amount of data for the machine learning model to identify the true trigger, which is crucial input when the ground truth is unknown and data difficult to obtain.\\
We assumed that all states (apparent and hidden) are independent and uniformly distributed. Whereas the equations would need to be modified if this assumption is invalid, the complexity of the challenge and the task to be solved by the machine learning model would be lesser. Hence, in this study we investigated and presented a solution to the most complex form of the challenge.
\\
We have not addressed a possible additional challenge: inaccuracy or inability in resolving the exact timing of data samples, which could result in variable (or wrong) ordering of elements in the sequence (e.g. LS vs SL). This can occur, for example, if the sampling frequency of the measurement or monitoring device is too low as compared to changes in the underlying process, or if a sampling buffer / queue is utilised. This additional challenge would inevitably increase complexity but not necessarily render a solution impossible. Rather, more data would be required with several potential triggers identified for further, iterative testing. Scenarios related to physical or computer processes may be more prone to this additional challenge than biological scenarios, which tend to take place at slower time scales or have a fixed sequence (e.g. DNA, for which, interestingly, sequencing errors could be treated as hidden states).\\  
We described and implemented a generally applicable algorithm capable of predicting the occurrence and identifying the trigger of an event using a deep learning model. Instead of using the results of the final output layer in the neural network, we suggested that the attention weight is a better interpreter for trigger identification. By combining multiple embedding methods, the model was enabled to utilise the distance between each element and event, the nearby elements of each element and the position of the element for trigger identification. Using the embedding layers together with an attention layer, the model was enabled to identify which elements contribute most to the event. Finally, the model returned the $k$ elements with the highest attention score, where $k$ is equivalent to the trigger length and, hence, reflects the top-ranked trigger candidates. With this information, the user is able to reconstruct and identify the trigger sequence that caused an event.\\
If the trigger length is unknown, $k$ can be used as an additional parameter in an iterative approach resulting in the model returning either a subset of the trigger, the entire trigger with some noise or the trigger if $k=l$. \\
In our machine learning experiments, we considered four scenarios, sequences with and without hidden states and with consecutive and non-consecutive triggers. For each scenario, we examined model performance using the minimum data set as calculated in Section \ref{min_data_subsec}. In the case of hidden states, high attention counts can appear for more than one trigger candidate. These trigger candidates could be tested one-by-one in an iterative approach, which is considerably more efficient than brute force testing of all possible triggers.  Further architectures and embedding methods for deep learning or different approaches (e.g. Bayesian inference) could be investigated too.\\
Importantly, we showed our derived equations to hold and found our proposed machine learning model capable of inferring the cause (trigger or trigger candidates) of the event and, therefore, inherently being able to predict events before they occur, despite crucial information being inaccessible.

\section{Conclusion}
We set out with an agent and their task to analyse a sequence of data points and predict and explain the occurrence of events in the face of uncertainties or hidden information. We formalised solutions in a combined analytical, simulation-based and machine learning approach, showed the derived equations to be valid and proved the developed algorithm to be able to predict events and infer causality despite incomplete information.\\
Our findings allow to assess the complexity of arbitrary scenarios and determine the required data to identify the triggers. When complexity increases, we recommend to combine the solutions presented here with an iterative approach to test which of the identified trigger candidates causes the event (e.g. by injecting trigger candidates). However, this approach may not be feasible in all domains.\\
The approach proposed in this paper can be used for root cause analysis, allowing to understand processes and subsequently implement improvements.

\section{Authors' Contributions}
YC, HL and SW derived the equations, analysed and interpreted the data and results, and wrote the manuscript. HL implemented the machine learning model. SW conceived and designed the study. NK, MC and AB contributed at various stages of the study. All authors reviewed the final manuscript. The authors thank Philippe Luc and Bill Jurasz for enjoyable discussions around this challenge. 

\newpage

\section{Appendix}

\begin{figure}[hb]
    \centering
    \includegraphics[width=.7\textwidth]{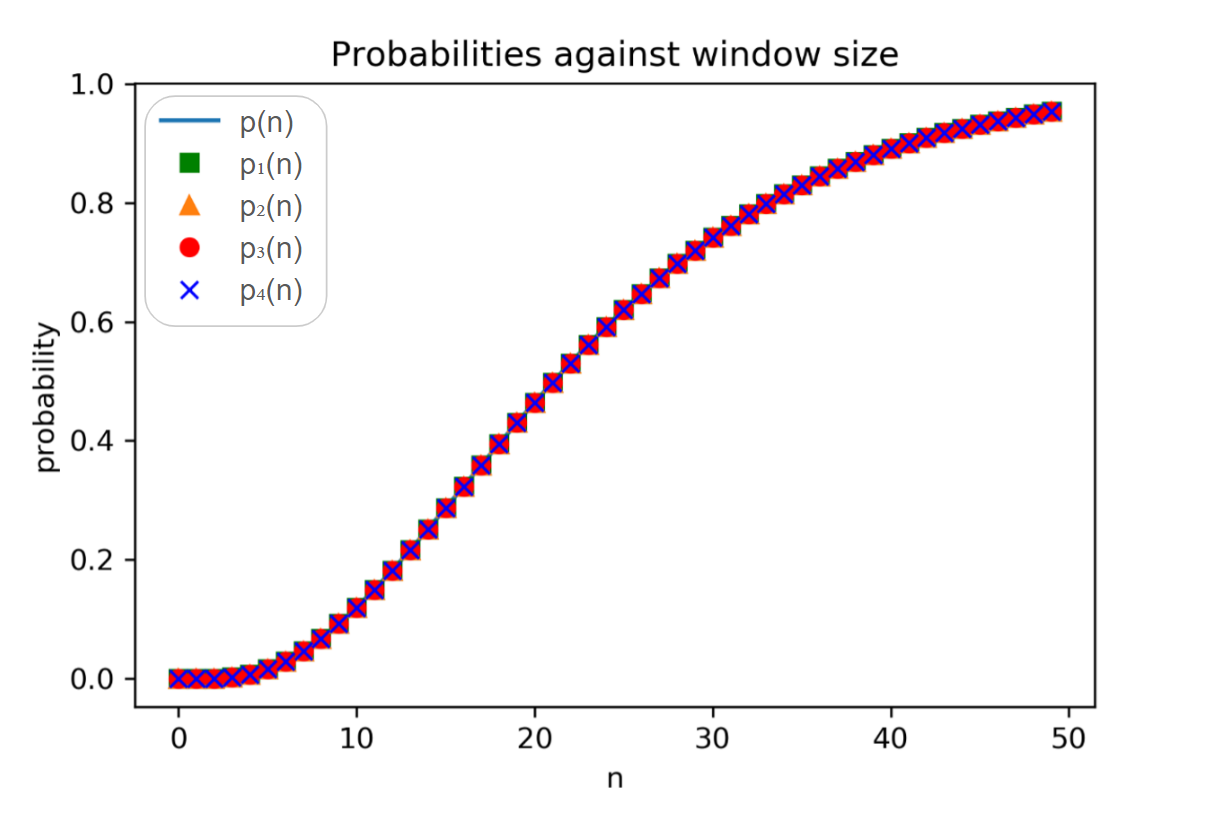}
    \caption{Simulation results demonstrating equivalence of all five equations across window length $n$.}
    \label{Different probability equation.}
\end{figure}

Here we show the equivalence of equation $p(n)$, a different form of $p(n)$ we refer to as $p_1(n)$ (please see below), and equations $p_{2-4}(n)$ introduced in Table \ref{table:gen}. These equations reflect different approaches to the same underlying challenge. We decided to include all approaches in this paper as for various readers some may be more intuitive than others. \\
The reader may be convinced of the equivalence of all five equations by inspecting Figure \ref{Different probability equation.}, which shows the outcome of computing the probabilities across different window lengths. However, the reader may want to read on for further informal proofs.
\subsection{$p_1(n)$}
\begin{figure}[b]
    \centering
    \includegraphics[width=.9\textwidth]{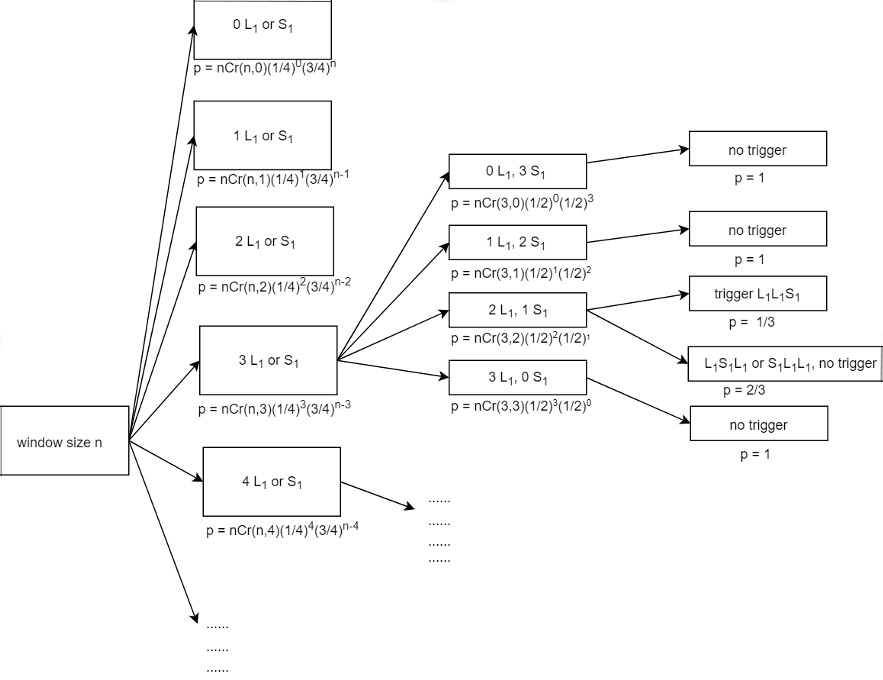}
    \caption{The tree diagram to illustrate $p_{1}(n)$}
    \label{tree1}
\end{figure}

Starting from Equation \ref{eq:1.1} for $p(n)$, we introduce $p_1(n)$: \\
\begin{equation}\label{eq:1.1}
    p_1(n)=\sum_{k=3}^n\binom{n}{k}\left(\frac{3}{4}\right)^{n-k}\left(\frac{1}{4}\right)^k\left[1-\sum_{x=0}^{x=k}\gamma(x,k)\frac{(y+1)\times x!\,y!}{(x+y)!}\right]
\end{equation}
Where $\gamma(x,k)$ is:
\begin{equation}\label{eq:gamma}
    \gamma(x,k)=\binom{k}{x}\left(\frac{x}{2}\right)^x\left(\frac{1}{2}\right)^{k-x}
\end{equation}\\
\textit{Informal Proof $p_1(n)$}: \\
We provide a proof here for the case of $L_iL_iS_i$, where $i\in H=\{1,2,3,4\}$. For demonstration purpose, we let $i=1$ here.\\
The first term $\binom{n}{k}\left(\frac{3}{4}\right)^{n-k}\left(\frac{1}{4}\right)^k$, represents the probability that exactly \textit{k} 'necessary elements' are chosen in a window of length $n$.\\
For the second term of Equation \ref{eq:1.1}, let the number of occurrence of $L_1$ be $x$ and the number of occurrence of $S_1$ be $y$, and $k=x+y$. Then $\gamma(x,k)$ is the binomial probability of having exactly $x$ lots of $L_1$ and $y$ lots of $S_1$. \\
If $x>2$ and $y>1$, there are $y+1$ types of combinations that the trigger $L_1L_1S_1$ does not exist, when we do not allow $S_1$ to appear after two $L_1$:\\
\systeme{S_1S_1\dots S_1S_1L_1L_1\dots L_1L_1 @(1),
         S_1S_1\dots S_1L_1S_1L_1\dots L_1L_1 @(2),
         S_1S_1\dots L_1S_1S_1L_1\dots L_1L_1 @(3),
         \dots ,
         L_1S_1\dots S_1S_1L_1L_1\dots L_1L_1 @(y+1)}
\\ \\
The probability of any of the above happening is $\frac{x!\,y!}{(x+y)!}$. Hence the probability that the true trigger $L_1L_1S_1$ does not exist given we have $x$ lots of $L_1$ and $y$ lots of $S_1$ is $\frac{(y+1)\times x!\,y!}{(x+y)!}$. Note that if $x<2$ or $y<1$, $\frac{(y+1)\times x!\,y!}{(x+y)!}=1$. Therefore, the second term defines the probability that the true trigger $L_1L_1S_1$ exists given the number of necessary elements ($L_1$ or $S_1$) equals $k$. This process can be visualised using a tree diagram (Figure \ref{tree1}).
\subsection{$p_2(n)$}
\begin{figure}[h]
    \centering
    \includegraphics[width=.9\textwidth]{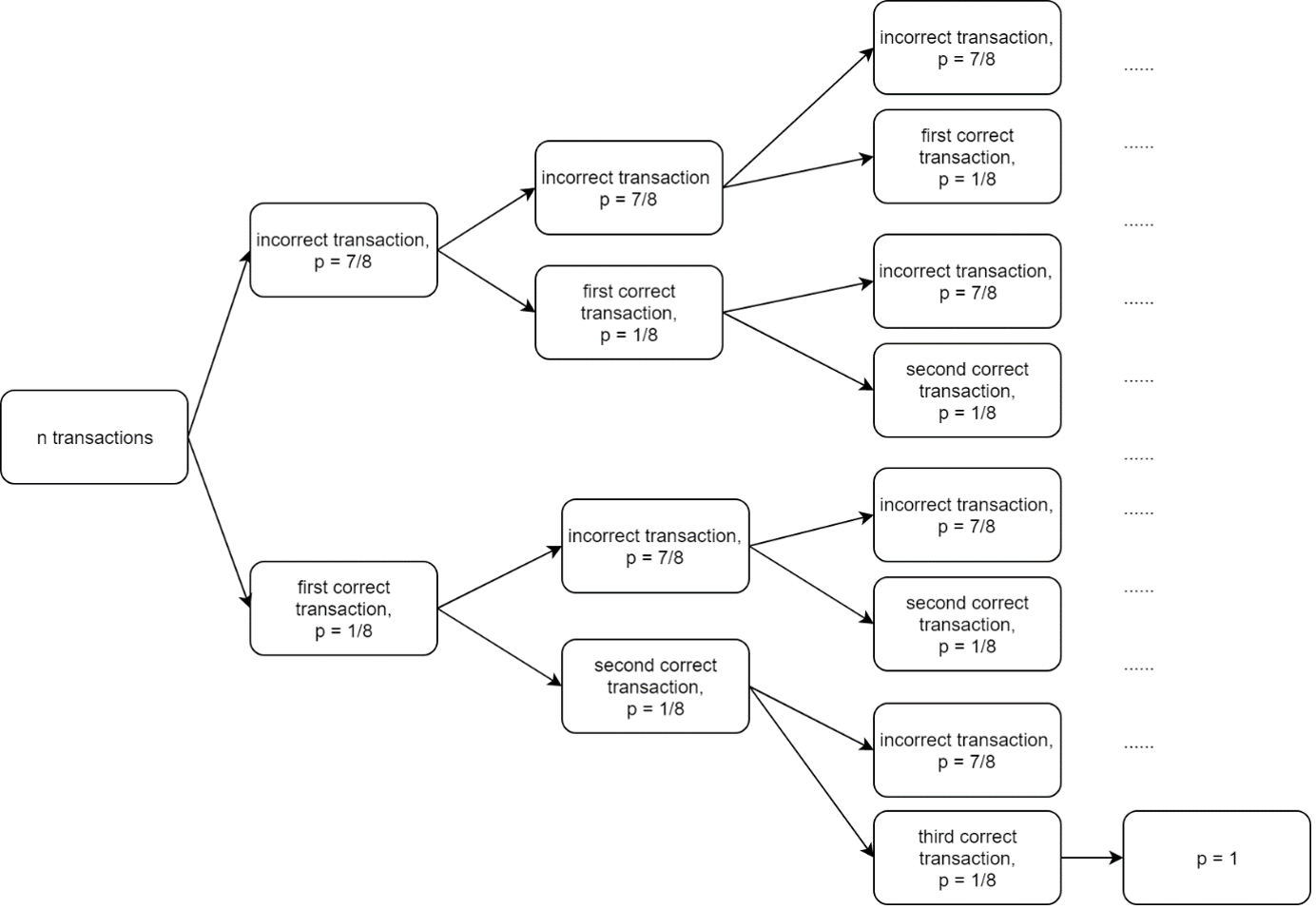}
    \caption{The tree diagram to illustrate $p_{2}(n)$}
    \label{tree2}
\end{figure}

Next, we address Equation $p_2(n)$ in Table \ref{table:gen}. Rather than choosing correct elements made of the trigger, one could construct $p{}(n)$ by exploring the probability trees of exactly one trigger is found in recurrence relations, which yields:
\begin{equation}\label{eq:1.2}
    p_2(n)=\sum_{i=3}^n\binom{i-1}{3-1}\left(\frac{7}{8}\right)^{i-3}\left(\frac{1}{8}\right)^3
\end{equation}
\textit{Informal Proof $p_2(n)$ :}\\
For every element there is a probability of $\frac{1}{8}$ that a 'correct element' can be chosen. For example, if the trigger is $L_1L_1S_1$ then the first correct element is $L_1$ with probability $\frac{1}{8}$, the same holds for the second and third element. This can be visualised in a tree diagram as shown in Figure \ref{tree2}, where each layer represents the next element in the sequence. Hence, the probability that the trigger occurs at the $j^{th}$ layer can be calculated by the probability that the $i^{th}$ layer has three correct elements multiplied by the probability that there are two correct elements in the previous $i-1$ layers.
\subsection{$p_3(n)$}
The probability can also be constructed in an iterative form, $p_3(n)$:
\begin{equation}\label{eq:1.3}
    p_{3}(n+1)=\frac{7}{8}p_{3}(n)+\frac{1}{8}\left(1-\left(\frac{7}{8}\right)^n-\left(\frac{7}{8}\right)^{n-1}\times n \times \frac{1}{8}\right)
\end{equation}
With initial condition: $p_3(0)=0$. \\ \\
\textit{Informal Proof $p_3(n)$ :}\\
The first term, $p_{3}(n)$, represents the probability that the $(n+1)^{th}$ element is not the last correct element for the trigger, and hence the true trigger has already occurred before the $n^{th}$ element. The second term, $\left(1-\left(\frac{7}{8}\right)^n-\left(\frac{7}{8}\right)^{n-1}\times n \times \frac{1}{8}\right)$, represents the probability when the $(n+1)^{th}$ element is the last correct element for the trigger, and there are at least two correct elements in the previous $n$ elements. Specifically, $\left(\frac{7}{8}\right)^n$ is the probability of no correct element has ever occurred in previous $n$ elements, and $\left(\frac{7}{8}\right)^{n-1}\times n \times \frac{1}{8}$ is the probability of only one correct element has occurred in the previous $n$ elements.
\subsection{$p_4(n)$}
Lastly, we address $p_4(n)$ in Table \ref{table:gen}. The recurrence relation can be further generalised and simplified by using the variable $l$ (the length of the true trigger):
\begin{equation}\label{eq:1.4}
    p_{4}(n+1,t)=\frac{7}{8}p_{4}(n,l)+\frac{1}{8}p_{4}(n,l-1)
\end{equation}
With initial conditions: $p_{4}(0,t)=0$, $p_{4}(n,0)=1$, and $p_{4}(n,n)=\left(\frac{1}{8}\right)^n$.\\ \\

\end{document}